\documentclass{article}

\usepackage{arxiv}

\usepackage[utf8]{inputenc} 
\usepackage[T1]{fontenc}    
\usepackage{hyperref}       
\usepackage{url}            
\usepackage{booktabs}       
\usepackage{amsfonts}       
\usepackage{nicefrac}       
\usepackage{microtype}      
\usepackage{graphicx}
\usepackage[authoryear, round]{natbib}
\usepackage{doi}
\usepackage{amsmath}

\title{GUT-IS: A Data-Driven Approach to Integrating Constructs and Their Relations in Information Systems\thanks{Accepted at the 34th European Conference on Information Systems (ECIS 2026), Milan, Italy.}}

\date{} 					

\author{
    Maximilian Reinhardt, Jonas Scharfenberger, Burkhardt Funk\\
	Institute of Information Systems\\
	Leuphana University\\
	Lüneburg, Germany\\
	\texttt{\{maximilian.reinhardt, jonas.scharfenberger, burkhardt.funk\}@leuphana.de}
}

\renewcommand{\headeright}{}
\renewcommand{\undertitle}{}
\renewcommand{\shorttitle}{GUT-IS}

\hypersetup{
pdftitle={GUT-IS: A Data-Driven Approach to Integrating Constructs and Their Relations in Information Systems},
pdfsubject={cs.CL, cs.LG},
pdfauthor={Maximilian Reinhardt, Jonas Scharfenberger, Burkhardt Funk},
pdfkeywords={Structural equation modeling, Construct clustering, Semantic similarity, Text embeddings},
}

\begin{document}
\maketitle

\setcounter{footnote}{0}

\begin{abstract}
Structural equation modeling is widely used in IS research. However, inconsistent construct definitions impede the cumulative development of knowledge. In this work, we present an approach that aims at the integration of structural equation models into a unified model: We use a combination of task-adapted text embeddings and clustering to produce a candidate set of construct groupings. Subsequently, we select the optimal solution using a loss function that explicitly trades off semantic purity and parsimony in the number of clusters. By making this trade-off explicit, our approach allows to analyze how construct groupings and their relations change as one shifts the priority from purity to parsimony. Empirically, we evaluate and explore the proposed methodology on two datasets from the IS domain.
\end{abstract}

\keywords{Structural equation modeling, Construct clustering, Semantic similarity, Text embeddings}

\section{Introduction}
Structural equation modeling (SEM) is one of the dominant empirical methods in IS research \citep{urbach_2010}. SEM-based studies have introduced and refined latent constructs to explain technology adoption, system success, digital work practices, and organizational transformation. Classic IS work has emphasized careful construct development and validation (see for instance \citeauthor{moore_1991}, \citeyear{moore_1991}). Yet, many reviews conclude that inconsistent construct definitions and measurement threaten validity and make it difficult to build a genuine cumulative tradition in IS \citep{li_2011}.

This tension is visible in what \citet{larsen_2016} term the construct jingle-jangle fallacy where conceptually similar constructs are labelled differently, or identical labels are used for distinct concepts, complicating literature reviews and meta-analysis. Despite sophisticated guidance on measurement models and construct specification (e.g.\ \citeauthor{petter_2007}, \citeyear{petter_2007}), much of the consolidation work around constructs is still performed ad hoc by individual reviewers. As a result, researchers face a large but fragmented landscape of constructs and relations that is difficult to synthesize.

At the same time, IS has started to develop knowledge infrastructures that encode theoretical content in machine-readable form. Platforms such as DISKNET \citep{dann_2019}, Promethis (\url{https://promethis.org}), and theorizeit (\url{https://theorizeit.org}) capture constructs, definitions, and estimated relations from SEM studies, and recent ontology-based frameworks propose semantic annotation and knowledge-synthesis pipelines to support large-scale, semi-automated literature reviews in IS \citep{huettemann_2024}. Parallel work on LLM-based extraction shows that constructs and relationships can be harvested from research articles at scale, albeit with non-trivial cleaning and validation effort \citep{scharfenberger_2025}.

We introduce GUT-IS – loosely inspired by Grand Unified Theories in physics \citep{langacker_1981} – as a data-driven pipeline that aims to integrate semantically similar constructs and reconcile their relations. In essence, we (1) start from existing IS knowledge repositories, (2) enrich construct names and definitions using LLM-based retrieval augmented generation, (3) learn a projection of domain-adapted embeddings, and (4) apply graph-based clustering governed by a loss function that trades off semantic purity against parsimony in the number of construct groups. This turns construct consolidation from a hidden, manual step into an explicit, tunable design decision and provides a unifying layer that can feed into meta-analytic preprocessing, ontology refinement, and ontology-based knowledge synthesis in IS.

\section{Related work}
Related work in this field encompasses construct and causal model identity detectors and tools that allow researchers to validate constructs against a set of existing constructs in the field of IS research. \citet{larsen_2016} propose a construct identity detection model that leverages latent semantic analysis (LSA) and algorithms that preserve word-order information to calculate text similarities between the items with which the constructs are measured. They evaluate their model on a dataset of 1,004 constructs which they manually assigned into clusters and report an ROC-AUC of 0.814. \citet{ludwig_2020} extend this work by considering the GloVe word embedding model. They also evaluate their model based on the ROC-AUC score on \citeauthor{larsen_2016}'s (\citeyear{larsen_2016}) dataset and report the best score (0.827) for using a weighted sum of GloVe embeddings of the words describing the construct items. 

\citet{song_2021} extend the construct identity problem to identifying similar causal models. They decompose the similarity measure into a structural component based on the graph edit distance and a semantic component which uses Wordnet to calculate similarities between construct names. \citet{larsen_2024} pursue a different goal and develop a tool that enables researchers to investigate how constructs load against each other and against a set of constructs published in IS research that are represented through 550 latent dimensions. To obtain this information, the user must provide their construct’s name, definition and items. The tool is built by combining fine-tuned word embeddings and principal component analysis; hyperparameters are determined using a human-in-the-loop approach.

In summary, related work presents useful tools to determine whether two constructs are identical and to enable researchers to compare constructs against a set of established ones. Thereby, \citet{larsen_2016} as well as \citet{ludwig_2020} rely solely on construct items, whereas \citet{larsen_2024} additionally utilize names and definitions. Causal relations between constructs are only used by \citet{song_2021} to calculate graph edit distances. In contrast, GUT-IS goes beyond the calculation of pairwise similarities and aims at the integration of constructs and their relations into a unified model. Further, we argue that there is not a single optimal grouping of constructs but that different groupings prioritize the parsimony and semantic purity of the resulting model differently. Our approach makes this trade-off explicit and allows to explore cluster solutions for different choices.

\section{Data and material} \label{sec:data}
We use two datasets to evaluate and explore our approach. First, we utilize the structural equation models stored in DISKNET \citep{dann_2019} and include 5,290 constructs from 578 publications for which full texts are available, as well as the 8,809 relations between these constructs. The information on constructs comprises names and definitions, however, with approximately 3,500 definitions missing. Relations are characterized by path coefficients and their statistical significance. The DISKNET data lacks ground truth on construct similarities, which is why we additionally use the data from \citet{larsen_2016} comprising 1,004 constructs. While this dataset does not contain relations, construct information includes manually determined assignments into 347 clusters. The cluster assignments imply pairwise similarities, from which 5,851 are positives and 497,655 negatives. Moreover, constructs are characterized by names, items, and definitions, with approximately 140 definitions missing.

We clean and enrich the construct names and definitions from both datasets using the LLM llama-3.3-70b-instruct\footnote{We gratefully acknowledge the use of the KISSKI API service SAIA \citep{doosthosseini_2026}, which provided access to llama-3.3-70b-instruct and e5-mistral-7b-instruct.}, which we choose as it is available as open source and competitive on common benchmarks \citep{meta_2024}. For the names, this includes expanding abbreviations to full English words and correcting spelling mistakes. Regarding definitions, the model is additionally prompted to ensure that they are meaningful and concise. Missing definitions are generated from scratch. To provide context for these tasks, we use a retrieval augmented generation approach on DISKNET \citep{dann_2019}: Besides the original name, definition, and publication title, the model is provided with chunks of the full texts which are retrieved using the original construct names as queries. For this purpose, we choose the text embedding model gte-Qwen2-1.5B-instruct \citep{li_2023}, as it offers a reasonable trade-off between performance and model size. For cleaning and enriching the dataset from \citet{larsen_2016}, we provide context in form of the original construct names, definitions, and items.

\section{Methods}
Our methodology consists of three core elements: (1) text embeddings for the calculation of pairwise similarities between constructs, (2) clustering methods for resolving logical inconsistencies in the resulting similarity graph, and (3) an interpretable loss function for the selection of optimal partitions.

\subsection{Pairwise similarity calculation}
We calculate similarities between constructs based on their names and definitions. While construct names alone can be ambiguous \citep{larsen_2016}, definitions specify the meaning of constructs in the context of the corresponding publications. Items might allow to capture even more fine-grained semantics, especially when constructs are similarly described but measured differently. However, their extraction is comparatively involving and difficult to automate. While definitions can be generated from context (see Section \ref{sec:data}), items need to be explicitly identified in publications. Therefore, we focus on names and definitions for scalability, although incorporating items might further increase performance.

Recent text embedding models can be used off-the-shelf to calculate similarities between textual inputs. However, as these models are typically trained on diverse text corpora, their representation may lack fine-grained similarities within specialized domains. One possible way to adapt pretrained text embeddings to specialized tasks is fine-tuning \citep{larsen_2024}. However, state-of-the-art models are large such that this requires extensive labels. Therefore, we propose to learn a lightweight projection that maps pretrained embeddings into a task-specific space. 

Specifically, we embed construct names and definitions separately using e5-mistral-7b-instruct \citep{wang_2024}, a recent model with a comparatively large capacity, and provide both embeddings as an input to the projection function. We parametrize the projection as a single linear neural network layer which we train by minimizing a contrastive loss \citep{hadsell_2006}, defined as
\begin{equation}
L(x_1, x_2, y) = 
\begin{cases}
1-\cos(x_1, x_2) & \text{if } y = 1 \\
\max(0, \cos(x_1, x_2)-m) & \text{if } y = -1
\end{cases}
\end{equation}
where $x_1$ and $x_2$ are embeddings, $y$ the true similarity between them, and $m \in [-1,1]$ a margin hyperparameter. Further, $\cos(\cdot,\cdot)$ denotes the cosine similarity and $\max(\cdot,\cdot)$ returns the maximum value. That is, the projection model is trained to bring the cosine similarity of similar pairs of instances closer to 1, and the cosine similarity of dissimilar pairs below the margin.

Because multi-cluster datasets yield heavily skewed similarity labels, we use hard negative sampling \citep{schroff_2015}. We define hard negatives as negative pairs with a cosine similarity above the margin, and semi-hard negatives as negative pairs which are below the margin but closer to it than a value $\delta$. All negatives below $m-\delta$ are considered easy. We sample positives and all negative types in fixed ratios to help the model improve on positives and hard negatives while maintaining its performance on easier cases.

\subsection{Resolving logical inconsistencies}
The learned embedding model can be used to calculate pairwise similarities for all constructs in a dataset, yielding a densely connected similarity graph. While such a graph might be insightful when looking at small groups of constructs, it will be hardly possible to identify a global structure. A first step towards a more interpretable similarity structure could be denoising the graph by discarding similarities below a threshold. However, the resulting graph will most probably contain logical inconsistencies: For example, construct A might be similar to construct B, and B to construct C, while the similarity between A and C falls below the threshold and is therefore removed. As means to resolve these logical inconsistencies and get a clear partitioning of constructs into disjoint sets, we consider three different clustering approaches that can be applied to the (denoised) similarity graph: Leiden method \citep{traag_2019}, spectral clustering \citep{luxburg_2007}, and agglomerative clustering \citep{jain_1999}.

\subsection{Navigating on the parsimony-purity continuum}
Forming groups of constructs based on pairwise similarities has several degrees of freedom: the threshold below which weak similarities are discarded, the choice of a clustering method, and the hyperparameters of the respective algorithm. We argue that the similarity of constructs is not simply positive or negative but inherently continuous, implying that there is not a single optimal combination of these choices. Instead, different configurations implicitly lead to a different balance between the parsimony and semantic purity of the resulting partition. Still, not every configuration might realize an optimum. There might exist multiple configurations which balance parsimony and purity similarly but yield a different partition quality. Therefore, we design an interpretable loss function that allows to explicitly trade off parsimony and purity and to identify the optimal configuration for a chosen balance. 

On a high level, we define our balanced loss function $L_{\text{balanced}}$ as a weighted combination of a parsimony loss $L_{\text{parsimony}}$ and a purity loss $L_{\text{purity}}$, controlled via a weighting parameter $\alpha$:
\begin{equation}
L_{\text{balanced}}(\alpha, C) = (1-\alpha)L_{\text{parsimony}}(C) + \alpha L_{\text{purity}}(C)
\end{equation}
where $C=\{C_1,C_2, \dots,C_k\}$ is a cluster solution for a set of constructs $X=\{x_1,x_2, \dots,x_n\}$ that partitions all constructs into disjoint sets. We define the parsimony loss as the ratio
\begin{equation}
L_{\text{parsimony}}(C) = \frac{|C|}{|X|} = \frac{k}{n}
\end{equation}
which lies in the interval $[1/n,1]$ and is minimized when all constructs are assigned to the same cluster.

For the purity loss, we propose two versions that focus on different analytical objectives: $L_{\text{relation\_purity}}$ focuses on optimizing the purity of the relations between clusters, while $L_{\text{construct\_purity}}$ aims at optimizing the purity of constructs within clusters. We define $L_{\text{relation\_purity}}$ as the weighted Shannon entropy of relations between constructs of distinct clusters. Let $R$ be the set of all unique relations between constructs of distinct clusters, and $R_{ij}$ the set of relations from $C_i$ to $C_j$. We calculate:
\begin{equation}
L_{\text{relation\_purity}}(C) = -\sum_{i \neq j}\frac{|R_{ij}|}{|R|}\Big(p_+(R_{ij})\log(p_+(R_{ij})) + p_-(R_{ij})\log(p_-(R_{ij}))\Big)
\end{equation}
where $p_+(R_{ij})$ and $p_-(R_{ij})$ are the proportions of positive and negative relations in $R_{ij}$, respectively. Note that the Shannon entropy takes values in $[0,1]$, bounding $L_{\text{relation\_purity}}$ accordingly. We handle the edge case where all constructs are in the same cluster by defining $L_{\text{relation\_purity}}=1$. Further, we define $L_{\text{construct\_purity}}$ as the average dissimilarity of a construct to all other constructs in the same cluster. Let $S_i$ be the set of all unique, undirected similarities between the constructs in cluster $C_i$. Similarities are normalized to $[0,1]$. We compute:
\begin{equation}
L_{\text{construct\_purity}}(C) = 1-\sum_{i:|S_i|>0}\frac{|C_i|}{|X|}\frac{\sum_{s \in S_i}s}{|S_i|}
\end{equation}
Only non-singleton clusters ($|S_i|>0$) contribute to the loss. If all clusters are singletons, we define $L_{\text{construct\_purity}}=0$, as this solution realizes maximal purity by definition. As similarities are normalized, $L_{\text{construct\_purity}}$ always lies between 0 and 1. Thus, all parsimony and purity loss terms take values in known intervals, enabling an interpretable trade-off via the parameter $\alpha$.

\section{Results}
The experiments consist of two parts: At first, we evaluate the proposed methodology on the labeled dataset from \citet{larsen_2016}. Second, we apply our approach to the DISKNET data \citep{dann_2019} to explore how clusters and their relations develop as we shift from parsimony to purity. Our source code can be found on GitHub (\url{https://github.com/MaxHReinhardt/gut-is}).

\subsection{Embedding model performance}
To evaluate the effectiveness of our similarity calculation, we conduct an ablation study with five distinct embedding versions: (1) the proposed approach, relying on embedded construct names and definitions which are projected into a joint task-specific space, (2, 3) projections of only names or definitions, testing the utility of inputs, and (4, 5) the original e5-mistral-7b-instruct embeddings of names or definitions, testing the effectiveness of the learned projection. 

We randomly split the constructs from \citeauthor{larsen_2016}'s (\citeyear{larsen_2016}) dataset into train (40\%), validation (20\%), and test (40\%) sets. We augment the train and validation data by generating paraphrased definitions using Meta's llama-3.3-70b-instruct \citep{meta_2024}. To find the optimal configurations for the learned projections, we conduct grid searches varying the embedding dimensionality, the margin hyperparameter, and the number of epochs to prevent overfitting. We retrain the projections using the best configurations ten times on the combined train and validation sets and evaluate their performances on the test set. The original embeddings are evaluated in a single run, as they are deterministic.

The results are shown in Table \ref{tab:roc}. The learned projection based on construct names and definitions achieves an ROC-AUC of 0.840 (std = 0.005) and thus outperforms the other tested approaches. It also surpasses the performance of the approaches presented by \citet{larsen_2016} and \citet{ludwig_2020}, who report ROC-AUC values of 0.814 and 0.827 on the same dataset. \citet{larsen_2024} obtain a higher ROC-AUC of 0.93 by fine-tuning text embedding models on a dataset of more than 6,000 constructs comprising their names, definitions and items. This performance gap indicates that fine-tuning might be superior to learning a projection when sufficient labeled data are available. Note that switching to fine-tuned embedding models can be done without changing the remaining parts of our workflow. The ablations show that removing either names or definitions leads to considerable performance drops. Furthermore, the ROC-AUC values of the original embeddings lie only slightly above 0.5, underlining the need for a task-specific approach.

\begin{table}[h]
\caption{Comparison of embedding approaches and text input versions. For the approaches with learned components, ROC-AUC values are reported as mean (standard deviation) across ten runs.}
\centering
\begin{tabular}{llc}
\toprule
\textbf{Embedding method} & \textbf{Text input} & \textbf{ROC-AUC} \\
\midrule
Learned projection & Names + definitions & \textbf{0.840 (0.005)} \\
Learned projection & Names & 0.793 (0.007) \\
Learned projection & Definitions & 0.765 (0.013) \\
Pretrained embeddings & Names & 0.544 \\
Pretrained embeddings & Definitions & 0.525 \\
\bottomrule
\end{tabular}
\label{tab:roc}
\end{table}

To assess the robustness of the pairwise similarity calculation towards central design choices, we additionally evaluate model performances under varying embedding models and hyperparameter values for the margin as well as the hard negative sampling parameter $\delta$ while keeping all other settings fixed. Besides e5-mistral-7b-instruct, we consider qwen3-embedding-8b \citep{zhang_2025} and llama-embed-nemotron-8b \citep{babakhin_2025} for generating pretrained embeddings, since all these models are in the range of 7-8 billion parameters and have an output dimensionality of 4096. The hyperparameters are varied across $m \in \{0.0,0.1, \dots, 0.5\}$ and $\delta \in \{0.00,0.05, \dots,0.20\}$. For each configuration, we train the projection model five times and measure the ROC-AUC on the test set.

\begin{figure}[h]
    \centering
    \includegraphics[width=0.9\textwidth]{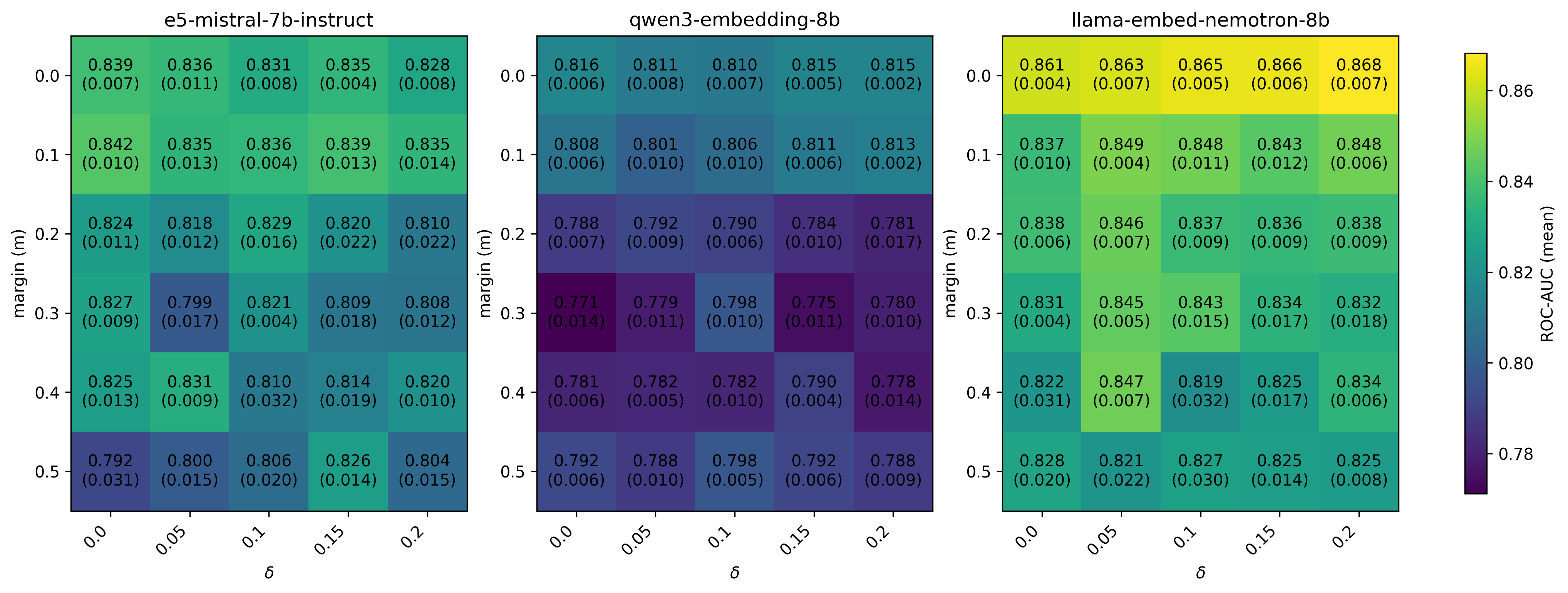}
    \caption{Robustness analysis of the pairwise similarity calculation, varying the embedding model, the margin ($m$), and the hard negative sampling hyperparameter $\delta$. ROC-AUC values are reported as mean (standard deviation) across five runs.}
    \label{fig:robustness_heatmap}
\end{figure}

Figure \ref{fig:robustness_heatmap} displays the results. The margin $m$ can have a considerable impact on performance, whereby low values generally tend to yield the best results. In comparison, $\delta$ appears to be of inferior importance, especially for low values of $m$. Changing the LLM can lead to relatively large performance differences with llama-embed-nemotron-8b achieving the best ROC-AUC of 0.868 (std = 0.007) and qwen3-embedding-8b falling short with a maximum score of 0.816 (std = 0.006). However, note that all inputs and parameters, except $m$ and $\delta$, are held fixed and that different LLMs might have different optima. With the optimal setup for each model, performance differences might diminish.

\subsection{Comparison of clustering approaches}
We compare agglomerative clustering, spectral clustering and the Leiden method across $\alpha$-values on the test split of the dataset from \citet{larsen_2016}. As input to the clustering algorithms, we calculate pairwise similarities between constructs using the learned embedding model. For each clustering method, we compute partitions using each hyperparameter combination in a search grid, varying both the threshold below which similarities are discarded, and either the number of clusters or the resolution parameter, depending on the method. For each $\alpha$ in $[0,1]$ with a step size of 0.05, we compute $L_{\text{balanced}}$ for each partition using $L_{\text{construct\_purity}}$ as the purity loss. Subsequently, we select the optimal partition for each combination of $\alpha$ and clustering method and compare it against the ground truth. The agreement of a cluster solution with the ground truth is measured using the adjusted mutual information (AMI) and the F1-score. As an additional reference, we compute the F1-score one achieves by binarizing the pairwise similarities from our embedding model using the optimal threshold.

\begin{figure}
    \centering
    \includegraphics[width=0.9\textwidth]{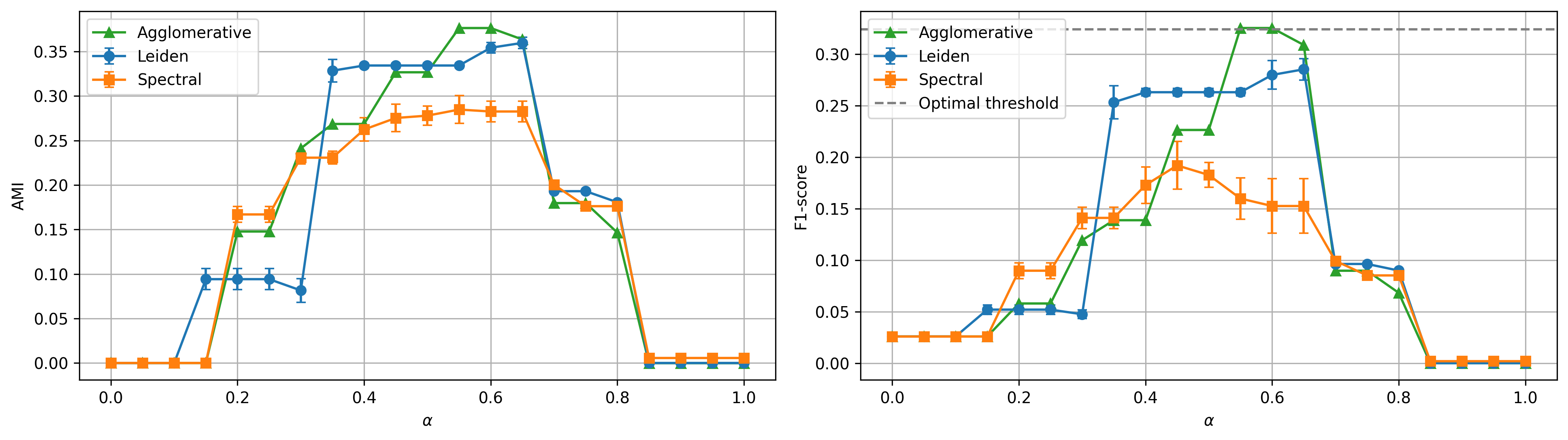}
    \caption{Comparison of clustering approaches across $\alpha$-values in terms of AMI (left) and F1-score (right). For clustering methods involving randomness, the plots show means and standard deviations across ten runs. The dotted line indicates the F1-score obtained by binarizing the embedding-based similarities using the optimal threshold.}
    \label{fig:cluster_comparison}
\end{figure}

Figure \ref{fig:cluster_comparison} visualizes the results. One can observe that all clustering methods are closest to the ground truth around an $\alpha$ of 0.6 in both metrics, except for the F1-score achieved with spectral clustering. This indicates that $\alpha=0.6$ balances parsimony and purity similarly as the true clusters. The highest agreement with the ground truth is achieved with agglomerative clustering. At the peak, the F1-score is at the same level as the score obtained by applying the optimal threshold to the pairwise similarities, showing that this clustering method can resolve logical inconsistencies in similarity relations without decreasing their quality. The Leiden method, while almost reaching a similar peak, diverges less smoothly from the ground truth when increasing or decreasing $\alpha$, and spectral clustering seems to be inferior across almost all $\alpha$-values. Thus, we conclude that agglomerative clustering is the most suitable choice for our methodological setup.

\subsection{Exploring the parsimony-purity continuum}
The exploration of cluster solutions across the parsimony-purity continuum can be approached from different angles, either with quantitative metrics aggregating core characteristics of the partitions, or qualitatively by analyzing the development of exemplary clusters, constructs, and relations. To avoid providing a selective view, we focus on a quantitative exploration in this short paper but strive to release a tool that allows for an interactive exploration soon. 

The exploration of the parsimony-purity continuum is carried out on the DISKNET data \citep{dann_2019}, since it contains theoretical relations. We compute similarities between all constructs using the embedding model trained on the dataset from \citet{larsen_2016}, which we assume to generalize well since both datasets contain constructs from the IS domain. Based on the resulting similarity graph, we perform agglomerative clustering for all combinations in a hyperparameter grid.

\begin{figure}
    \centering
    \includegraphics[width=0.9\textwidth]{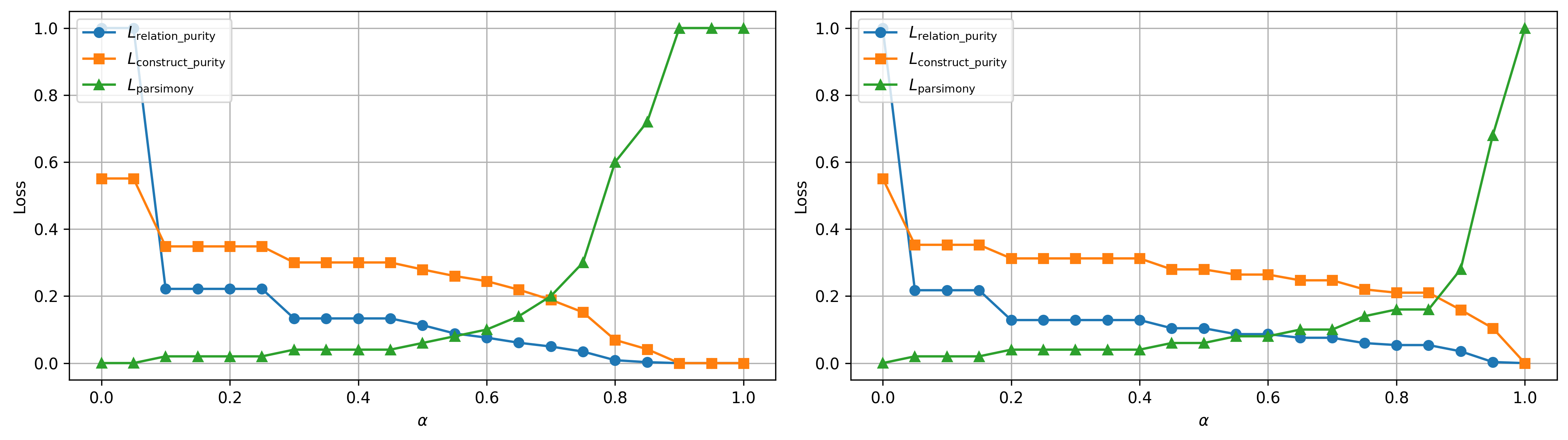}
    \caption{Development of parsimony and purity losses across $\alpha$-values. Optimal partitions are selected using either $L_{\text{construct\_purity}}$ (left) or $L_{\text{relation\_purity}}$ (right) as purity loss.}
    \label{fig:continuum_exploration}
\end{figure}

Figure \ref{fig:continuum_exploration} shows how parsimony and purity of the clustering solutions develop across $\alpha$-values, using either $L_{\text{relation\_purity}}$ or $L_{\text{construct\_purity}}$ as the purity loss to select the optimal partitions. In either case, both purity losses decrease and the parsimony loss increases with increasing $\alpha$, explicitly demonstrating the trade-off. While all losses exhibit strong increases as their weight goes to 0, especially the parsimony loss shows a highly non-linear development. Further, with either choice of purity loss, $L_{\text{construct\_purity}}$ and $L_{\text{relation\_purity}}$ exhibit a relatively strong correlation. We therefore conclude that, while focusing on different objectives, both losses contribute to the optimization of semantic purity in a broader sense.

\section{Discussion}
We present GUT-IS, a data-driven workflow which aims at integrating and unifying structural equation models from the IS domain. Conceptually, GUT-IS switches from calculating pairwise similarities to forming globally coherent similarity structures, which is achieved via clustering. Optimal cluster solutions are determined using a loss-based selection mechanism that balances the conflicting objectives of obtaining a parsimonious but semantically pure partition. GUT-IS makes this trade-off an explicit choice and thus provides a principled way to navigate on the continuum between parsimony and purity.

From a practical standpoint, we achieve a high partition quality while striving for minimal manual effort: Construct similarities are derived from construct names and definitions, which are cleaned and enriched using an LLM, and computed by combining a pretrained embedding model with a task-specific projection layer that can be trained with little labeled data. Still, our results show that the model performs competitively in predicting pairwise similarities, compared to existing approaches \citep{larsen_2016, ludwig_2020}. Further, we observe empirically that the application of clustering methods can resolve logical inconsistencies in pairwise similarities while maintaining their overall quality. Despite our experiments focus on data from the IS research, GUT-IS is portable to other domains.

\subsection{Limitations}
A first limitation of our work is the limited evaluation and exploration on the DISKNET data \citep{dann_2019}. While we could validate most parts of our methodology using the labeled dataset from \citet{larsen_2016}, this was not possible for the retrieval augmented generation of construct names and definitions, since the dataset does not link constructs to publications. Positive indication can be drawn from the LLM-based data cleaning and enrichment on \citeauthor{larsen_2016}'s (\citeyear{larsen_2016}) dataset which contributes to the successful calculation of similarities; furthermore, the chosen embedding model performs competitively on retrieval benchmarks \citep{li_2023}. However, this step currently remains without explicit verification. In addition, the exploration of the parsimony-purity continuum is purely quantitative and lacks a qualitative perspective. Another limitation of our results is the training of the projection model based on the cluster-based gold labels from \citet{larsen_2016}. While the clusters are aimed at containing correspondent constructs, they do not necessarily reflect exact semantic equivalences, which might introduce label noise and limit the validity of the similarity calculation.

\subsection{Future work}
To enable a convenient and efficient qualitative analysis, as a next step, we currently develop an interactive tool that allows to freely choose a balance between parsimony and purity and to analyze the resulting clusters as well as their relations. We intend to present the tool at the ECIS conference and subsequently enable its public use by other IS researchers.

\bibliographystyle{abbrvnat}
\bibliography{references}

\end{document}